\documentclass[conference]{IEEEtran}
\IEEEoverridecommandlockouts
\usepackage{amsfonts}
\usepackage{textcomp}
\usepackage{multirow}
\usepackage{amsmath}
\usepackage{algorithm}    
\usepackage{algorithmic}   
\usepackage{cite}
\usepackage{verbatim}
\usepackage{booktabs} 
\usepackage{color}
\usepackage{subfigure}
\usepackage{graphicx} 
\usepackage[colorlinks,linkcolor=blue]{hyperref}

\usepackage{amsmath,amssymb,amsfonts}
\usepackage{algorithmic}
\usepackage{graphicx}
\usepackage{textcomp}
\usepackage{multirow}
\usepackage{array}
\usepackage{subfigure}
\usepackage{booktabs}
\usepackage{amsmath}
\usepackage{algorithm}    
\usepackage{algorithmic}   
\usepackage{wrapfig}

\def\BibTeX{{\rm B\kern-.05em{\sc i\kern-.025em b}\kern-.08em
    T\kern-.1667em\lower.7ex\hbox{E}\kern-.125emX}}

\def\aboveeq{\setlength{\abovedisplayskip}{6pt}}
\def\downeq{\setlength{\belowdisplayskip}{6pt}}

\begin{document}

\title{Optical Flow Based Online Moving Foreground Analysis\\
{\footnotesize \textsuperscript{*}}
\thanks{}
}

\author{

\IEEEauthorblockN{1\textsuperscript{st} Junjie Huang}
\IEEEauthorblockA{\textit{Institute of Automation} \\
\textit{Chinese Academy of Sciences}\\
Beijing, China \\
huangjunjie2016@ia.ac.cn}
\and
\IEEEauthorblockN{2\textsuperscript{nd} Wei Zou}
\IEEEauthorblockA{\textit{Institute of Automation} \\
\textit{Chinese Academy of Sciences}\\
Beijing, China \\
wei.zou@ia.ac.cn}
\and

\IEEEauthorblockN{3\textsuperscript{rd} Zheng Zhu}
\IEEEauthorblockA{\textit{Institute of Automation} \\
\textit{Chinese Academy of Sciences}\\
Beijing, China \\
zhuzheng2014@ia.ac.cn}
\and
\IEEEauthorblockN{3\textsuperscript{rd} Jiagang Zhu}
\IEEEauthorblockA{\textit{Institute of Automation} \\
\textit{Chinese Academy of Sciences}\\
Beijing, China \\
zhujiagang2015@ia.ac.cn}

}

\maketitle

\begin{abstract}
Obtained by moving object detection, the foreground mask result is unshaped and can not be directly used in most subsequent processes. In this paper, we focus on this problem and address it by constructing an optical flow based moving foreground analysis framework. During the processing procedure, the foreground masks are analyzed and segmented through two complementary clustering algorithms. As a result, we obtain the instance-level information like the number, location and size of moving objects. The experimental result show that our method adapts itself to the problem and performs well enough for practical applications.
\end{abstract}


\section{INTRODUCTION}

The detection of moving object is popular for researching. Aiming at detecting moving objects from complex scenes, many methods have been proposed \cite{DSMOD,OBSMC,MODBOS,FST,MCD,RFMOD}. Besides, some datasets have been collected and published \cite{SMOLTVA,CDNet2014}, promoting the development of moving object detection. Among these works, a dominant paradigm for the output of moving object detection is some foreground masks. These masks consist of pixel-wise labels which provide a detailed discriminative result indicating whether a pixel belong to the moving foreground. But for the oncoming issue like track and instance analysis, it lacks of some result with practical value, as there aren't any direct outputs indicating how many moving objects in the scenes, where are them, what are the sizes of them and which pixels belong to the same moving object, et.al. Compared to the pixel-wise labels, these outputs are more helpful for the subsequent processes. Thus, to a certain extent, the foreground masks obtained by aforementioned methods are unshaped and postprocessing procedures are needed for obtaining the instance-level information of moving objects.

We address the problem of moving foreground analysis by constructing an optical flow based framework. Optical flow is used as the main feature of pixels providing crucial information during the analysis processes. Optical flow has been extensively studied \cite{FlowField,SegFlow,OFMC,FlowNet2,TVL}, and has been widely used for video analysis \cite{SMOLTVA,HOF,HBPCFOF,BMOD} benefited from its direct reflection of the scene's motion information. However, to obtain the satisfactory result, researchers still encounter many obstacles because of some common problems. Such as precision problem caused by large displacements, occlusion and intensity changes \cite{OFMC}, computation cost problem caused by algorithm complexity \cite{OFMC}, and application problem caused by the discrete distribution of optical flow. FlowNet2.0 proposed in \cite{FlowNet2} offers an optical flow estimation that is accurate enough for practical application and fast enough for online application. 

\begin{figure}[t]
\begin{center}
\includegraphics[width=7.5cm]{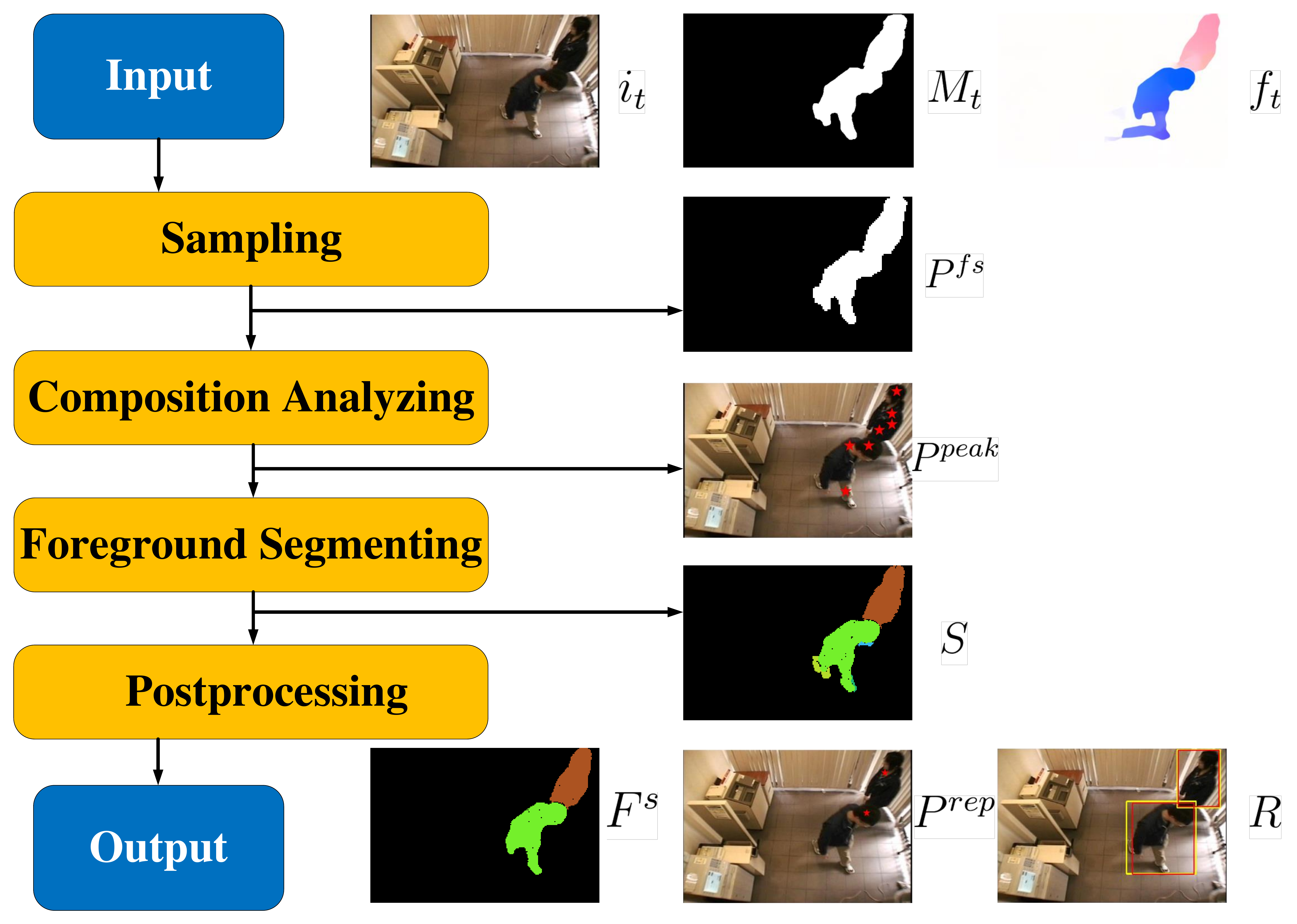}
\end{center}
\caption{Visualization of the moving foreground analysis framework.}
\label{fig1}
\end{figure}

In many other works like \cite{FST,SMOLTVA}, optical flow was estimated between adjacent frames. However, under most situations, due to high frame rate and relatively low object moving speed and unsteady motion, the optical flow distinction between foreground and background is too small to be directly used to extract the foreground targets correctly under the influence of interference. To enhance the feature discrimination between foreground and background, a dominant paradigm is using point trajectories as the feature like \cite{SMOLTVA}. As it needs a point tracking procedure to obtain the point trajectories, the shortcomings of this method include additional consumption of computing and storage resource, introduction of extra interference and higher complexity of feature. As described in Section~\ref{sec:OE}, we adopt an approximation of point trajectories to effectively enhance the feature discrimination while avoiding these problem.

Our framework integrates the foreground mask, the optical flow and the intensity information to find out some useful instance-level information. To this end, as shown Figure~\ref{fig1}, there are two major steps: composition analysis and foreground segmentation. Composition analysis addresses the problems of how many moving objects in the scene and which points can be used to initially locate them. Partitioning clustering algorithms are qualified for these problems as they produce single-level clustering result \cite{CCA}. In this work, we adopt Clustering by Fast Search and Find of Density Peaks (CFSFDP) \cite{CFSFDP} method as it provides an automatic mechanism for analyzing the number of clusters and offering a representative center for each cluster, which can meet our demands. On the other hand, foreground segmentation addresses the problem of differentially labeling pixels between different instances. Hierarchical clustering algorithms fit this problem as they output multi-level nested decompositions \cite{CCA}. Considering the irregular moving object shape and the continuous optical flow distribution inside a moving object, Graph-Based Image Segmentation(GBIS) \cite{GBIS} method is applied in this step as it can be performed efficiently and outputs suitable result. We combine the result obtained from these two steps by using the result of composition analysis to guide the foreground segmentation and using the result of foreground segmentation to merge the result from composition analysis. As a result, we suppress most false positive result and obtain a high precision. Furthermore, two situations are considered within our procedure. First is that the two instances own different movement information. And the second is that the two instances are spatially apart from each other. These provide guidance for procedure design and give expression in the processing procedure.

We evaluate the proposed moving foreground analysis framework on ten video sequences provided by ChangeDetection2014 dataset(CDnet2014) \cite{CDNet2014}. Qualitative result and quantitative result are both list in Section~\ref{sec:CC} for reference. Moreover, the complementary effect of the two clustering algorithms is explored to proof the necessity. We also test the efficiency of the proposed framework in depth and offer some advice for practical applications.

The contributions of this work are as follows: Firstly, we construct an efficient optical flow based framework for addressing the problem that analyzes the foreground masks obtained by moving object detection, and output instance-level information with practical value. Secondly, through experiment, we demonstrate that the proposed framework adapts itself to the problem and offer some advises for applications.

The remainder of this paper is organized as follows. Our proposed analysis framework based on optical flow is introduced in Section~\ref{sec:MT} and its effectiveness is verified in Section~\ref{sec:EP} by comprehensive experiments. Finally, Section~\ref{sec:CC} is devoted to conclusions.


\section{Methodology}
\label{sec:MT}
Our goal is to segment the moving object detection result into different instances and obtain instance-level information of moving objects. To this end, we construct the processing framework as shown in Figure~\ref{fig1}. There are mainly four processes: sampling, composition analyzing, foreground segmenting and postprocessing. In the following, each part of the framework is introduced in detail.

\subsection{Optical Flow Estimating}
\label{sec:OE}
We adopt an approximation of point trajectories, that is estimating optical flow between frame $t$ and frame $t-k$, where $k$ is an integral parameter related to the application context and limited by the ability of optical flow estimation algorithm calculating large displacement. As described in Formula~\eqref{eq1}, vector sums of $k$ optical flow vectors is used to replace the point trajectories in \cite{SMOLTVA}. The feature discrimination can be enhanced in a reasonable way which is quite simple but efficient.
{\aboveeq
\downeq
\begin{equation}
\label{eq1}
 \textbf f_{t,t-k}=\textbf f_{t,t-1}+\textbf f_{t-1,t-2}+\dots+\textbf f_{t-k+1,t-k}
\end{equation}}
where $\textbf f_{t,t-k}={
\left[ \begin{array}{cc}
u&v
\end{array} 
\right ]}^T$ is the optical flow vector which projects a 2D location $\textbf p_t$ in frame $t$ to the location $\textbf p_{t-k}$ in specified frame $t-k$.

\begin{figure}[htbp]
\centering
\includegraphics[width=0.5\textwidth]{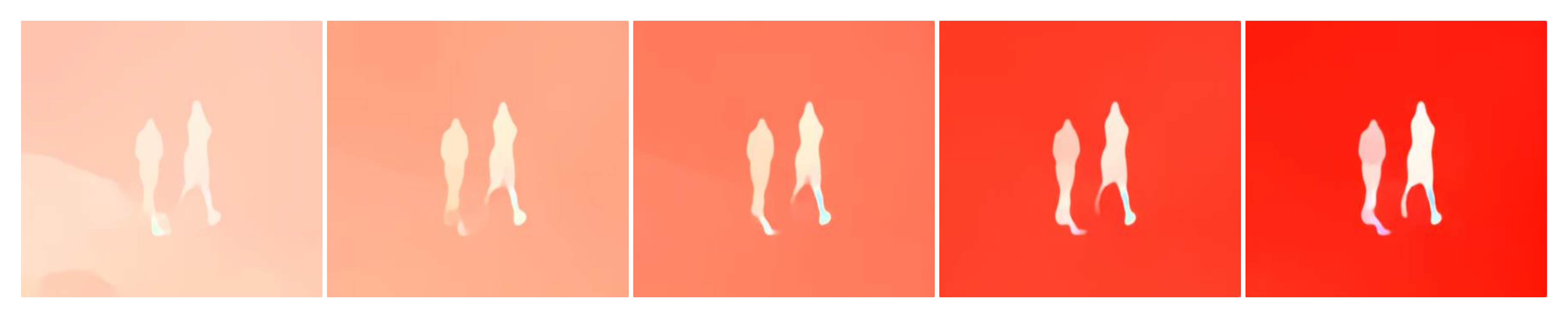}
\caption{Optical flow field of different $k$ value: from left to right $k=1$, $k=2$, $k=3$, $k=4$, $k=5$. The hue of a pixel indicates the optical flow direction and the color saturation indicates velocity. }
\label{fig2}
\end{figure}
As shown in Figure~\ref{fig2}, with the increment of $k$, the difference between foreground and background gradually becomes evident, which has a significant impact on the moving object detection. Taking into account speed and accuracy, FlowNet2.0 \cite{FlowNet2} is used to compute the optical flow vectors, which are points from the latest frame $t$ to frame $t-k$. The optical flow vectors are used as the main feature in the following procedures.

\subsection{Sampling}\label{CCA}

We sparsely sample points and only retain 1/9 of the total points as sampling too much points contributes little to improve the system's performance, but causes a huge amount of computation. In this paper, the computational complexity of using method CFSFDP \cite{CFSFDP} is proportional to the square of the number of sample points. And the computational complexity of using method GBIS \cite{GBIS} represents linearity correlation to the number of sample points. In addition, we find out the foreground sample points $P^{fs}$ from the all sample points $P^s$ utilizing the foreground masks $M_t$ provided by ground truth:
{\aboveeq
\downeq
\begin{equation}
\label{eq9}
P^{fs}=P^s \cap M_t 
\end{equation}}
\subsection{Composition analyzing}
\label{CCB}
\begin{figure}[htbp]
	\centering
	\includegraphics[width=0.48\textwidth]{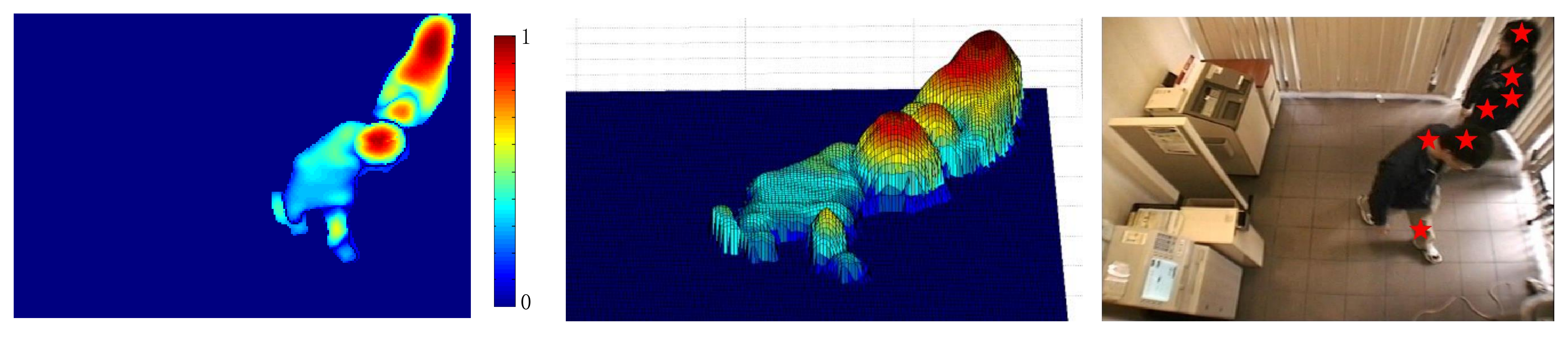}
	\caption{Density map of sample points. A sample image from CopyMachine video \cite{CDNet2014}. Some major peaks are shown in the last image drew in red stars.}
	\label{fig3}
\end{figure}

In this subsection, we aim at finding out how many moving objects in the scene and initially locating them by using some representative points. The density map's peak points defined in CFSFDP \cite{CFSFDP} method reflect different individuals as shown in Figure~\ref{fig3}. 
To find out the peak points, the feature of sample point $i$ is defined as: 
{\aboveeq
\downeq
\begin{equation}
	F_i=[\begin{array}{cccc}
		u_i & v_i& x_i/p & y_i/p
	\end{array}]^T
\end{equation}}
where $x^i$ and $y^i$ are the coordinates of the sample point $i$. $p$ is a parameter used to trade off the influence from optical flow and coordinates. A random sampling is adopted to maintain $N_c\leq200$ points, for the sake of controlling the computational consumption of CFSFDP below a certain level while making as little influence as possible to the result. Then the density of sample point $i$ is defined as:
{\aboveeq
\downeq
\begin{equation}
	\rho_i=\sum_{j=1}^{N_c}e^{-(\frac{d_{ij}}{d_c})^2}
\end{equation}}
where $d_{ij}=|| F_i-F_j||_2$ is the distance between two points in a data space, and $d_c$ is a threshold. After that, the minimum $d_{ij}$ between one point $i$ and the other point $j$ which has higher density is used to defined $\delta_i$ as:
{\aboveeq
\downeq
\begin{equation}
	\delta_i^w={\begin{cases}
		\min \limits_{j:\rho_j>\rho_i} ( d_{ij}^w ) ,\;  & \exists \rho_j>\rho_i\\
		\max \limits_{j=1,2,\cdots,N_c} ( d_{ij}^w),\; & \exists \rho_j>\rho_i
	\end{cases}},w=f, c
\end{equation}}
where 
\begin{align*}
	&d_{ij}^f = || F_i^f-F_j^f||_2,\; F^f={
	\left[ \begin{array}{cc}
		u_{t,t-k} & v_{t,t-k}
	\end{array} 
	\right ]}^T\\
	&d_{ij}^c = || F_i^c-F_j^c||_2,\; F^c={
	\left[ \begin{array}{cc}
		x_t & y_t
	\end{array} 
	\right ]}^T
\end{align*}
In this work, we define two kinds of $\delta_i$: $\delta_i^f$ and $\delta_i^c$. $\delta_i^f$ is calculated using optical flow, and $\delta_i^c$ is coordinates. Finally, the peak points $ P_{peak}$ are judged out by the following criterion:
\begin{equation}
	 P^{peak} =\{ P^{fs}_i\mid \rho_i>T_{r}\&\&(\delta^{f}_i>{T_{d1}} \| \delta^{c}_i>{T_{d2}})\}
\end{equation}
where $T_r=\rho_{max}/c_1$, $T_{d1}=c_2*k$ and $T_{d2}$ are three thresholds. The condition $\delta^{f}_i>{T_{d1}} \| \delta^{c}_i>{T_{d2}}$ means that a peak point should own different optical flow compared to the points with higher density, or spatially apart from them. Condition $\rho_i>T_{r}$ is used to exclude the outliers. $T_r$ and $T_{d1}$ are set as adaptive thresholds, where $\rho_{max}$ is the maximum density inside the current frame.

\subsection{Foreground segmenting}\label{CCC}
In this part,we adopt the GBIS \cite{GBIS} method to divide the foreground points into different sections. Optical flow is used as the feature of each pixel. Different from Felzenszwalb, et.al \cite{GBIS}, under the influence of sparse sampling, we construct the edges between points and their eight nearest neighbors to ensure the continuity of an instance. However, this will produce many superfluous edges as three edges for a point are enough for ensuring the connectivity of the graphs. So, the four edges with minimum edge weight are constructed in practice. Then, Algorithm 1 in GBIS \cite{GBIS} is used to segment the foreground. In Algorithm 1, the parameter $\tau$ is used to control the degree of polymerization in the form of $\tau/C$, where $C$ is the number of points inside a group. $\tau$ reflects the desired size of output groups, we set it as an adaptive variable indicating the desired object size:
\begin{equation}
 \tau=\frac{2*N_{P^{fs}}}{N_{P^{peak}}}
\end{equation}
where $N_{P^{fs}}$ denotes the number of foreground sample points. $N_{P^{peak}}$ denotes the number of peaks obtained in Section~\ref{CCB}. After finishing Algorithm 1, a set of segment result $S=\{S_j\}$ is obtained.

\subsection{Postprocessing}
Firstly, the section $S_j$ that includes any peak points is selected to construct a set of final foreground instances :

\begin{equation}
\label{eq14}
F^s=\{ S_j \mid \exists P^{peak}_{i} \in S_j \}. 
\end{equation}

Secondly, the peaks obtained in section~\ref{CCB} are filtered. Specifically speaking, among all the peaks that belong to the same section, only the one with highest density is retained on behalf of this section and a set of representative peaks is obtained as:

\begin{equation}
\label{eq15}
P^{rep}=\{P^{peak}_j \mid \forall P^{peak}_{i} \in F^s_j:\rho_j \geq \rho_i\}
\end{equation}

At last, a minimum bounding box $R^{s}_{i}$ is used to include all sample points that belong to the same foreground $F^{s}_{i}$. Then we obtain a bounding box $R_i$ corresponding to a moving foreground $F_{i}$ by slightly enlarging $R^{s}_{i}$. In practice, we enlarge the width and height of the bounding box $R^{s}_{i}$ by a specific pixel number which is equal to the sparse sample interval.

\begin{algorithm}[htbp]
	\caption{Moving Foreground Analysis}
	\begin{algorithmic}[1]
	\STATE\textbf{Input:} image sequence $I_t$, foreground masks $M_t$, optical flow $\textbf f_{t,t-k}$;
	\STATE sampling $I_t$ to obtain sample points $P^s$;
	\STATE utilizing $M_t$ to judge out foreground sample points $P^{fs}$ by Formula~\eqref{eq9};
	\STATE performing CFSFDP method to find out the peak points $P^{peak} $ from $P^{fs}$;
	\STATE performing Algorithm 1 in \cite{GBIS} on $P^{fs}$ to produce segments $S$;
	\STATE selecting final foreground segment $F^s$ by Formula~\eqref{eq14};
 	\STATE selecting final representative peak points $P^{rep}$ by Formula~\eqref{eq15};
	\STATE estimating a bounding box $R_i$ for every foreground segment in $F^s$;
	\STATE\textbf{Output:} representative peak point $P^{rep}_{i}$, foreground segment $F^s_i$ and a bounding box $R_i$ of moving object $i$ 
    \end{algorithmic}  
\end{algorithm}

\begin{figure*}[htbp]
	\centering
	\includegraphics[width=0.95\textwidth]{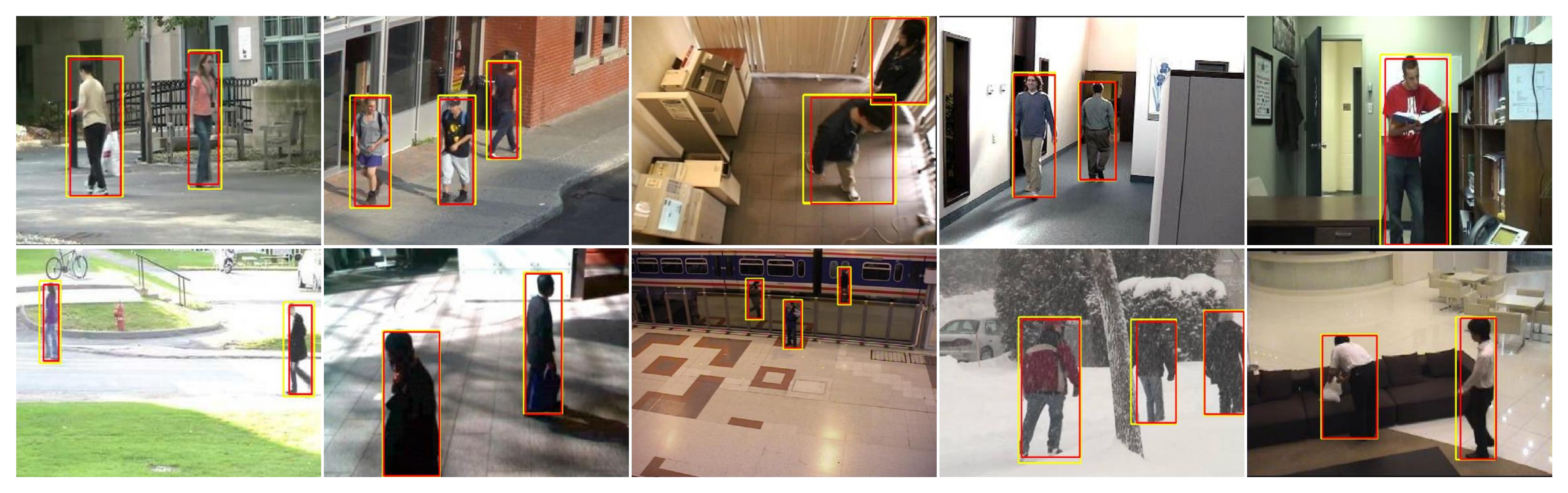}
	\caption{The bounding box results in some key frames. The yellow bounding boxes denote the ground truth and the red bounding boxes denote the result of our method.}
	\label{fig4}
\end{figure*}

\section{Experiment}
\label{sec:EP}
The proposed method is implemented with Matlab, and is tested with ten video sequences that contain different numbers of foreground instances in different scenes. In this section, the video sequences and evaluation metrics are introduced first. Then, we test the proposed method's output qualitatively and quantitatively. Finally, the contribution of composition analysis and the frame rate of the proposed method are explored respectively.

\subsection{Dataset}
The video sequences are provided by the pedestrian subset of ChangeDetection2014 dataset (CDnet2014) \cite{CDNet2014}. CDnet2014 offers three vital needed information in our experiments: intensity images of video sequences and pixel-wise foreground masks for input, instance-level annotation in the form of bounding boxes for output reference. The pedestrian subset contains ten video sequences and total 16864 bounding boxes for pedestrian annotation. Foreground analysis in these video is challenge as the scenes contain uncertain numbers, unbalance sizes and irregular shapes of objects. Besides, occlusion problem is another main challenge. 

\subsection{Evaluation Metrics}\label{AA}
In this section, we discuss the main three different metrics for the method performance evaluation: Intersection over Union(IoU), Recall(Re) and Precision(Pr). IoU metric is introduced to measure the accuracy of the bounding box result. It also used as a threshold for judging whether an instance is correctly analyzed. Recall (Re) metric is used to reflect how well the method figure out the instance level information. As shown in Figure~\ref{fig6}. The methods whose recall curves are close to the top and right of the plot have high success rate and quality respectively. Precision (Pr) metric is used to reflect how many mistake made by methods make. As shown in Figure~\ref{fig6}. The methods whose precision curves are close to the top of the plot produce little false positive results.

\subsection{Parameter setting}\label{AA}
We estimate optical flow between $I_t$ and $I_{t-k},\; k=5$. In the sampling process, the sample interval was set as $s=3$. For CFSFDP, the balanced parameter was set as $p=50$, and the thresholds were set as $c_1=15$, $c_2=0.5$ and $T_{d2}=50$.

\subsection{Quantitative result}\label{AA}
\begin{figure}[h]
	\centering
	\includegraphics[scale=0.50]{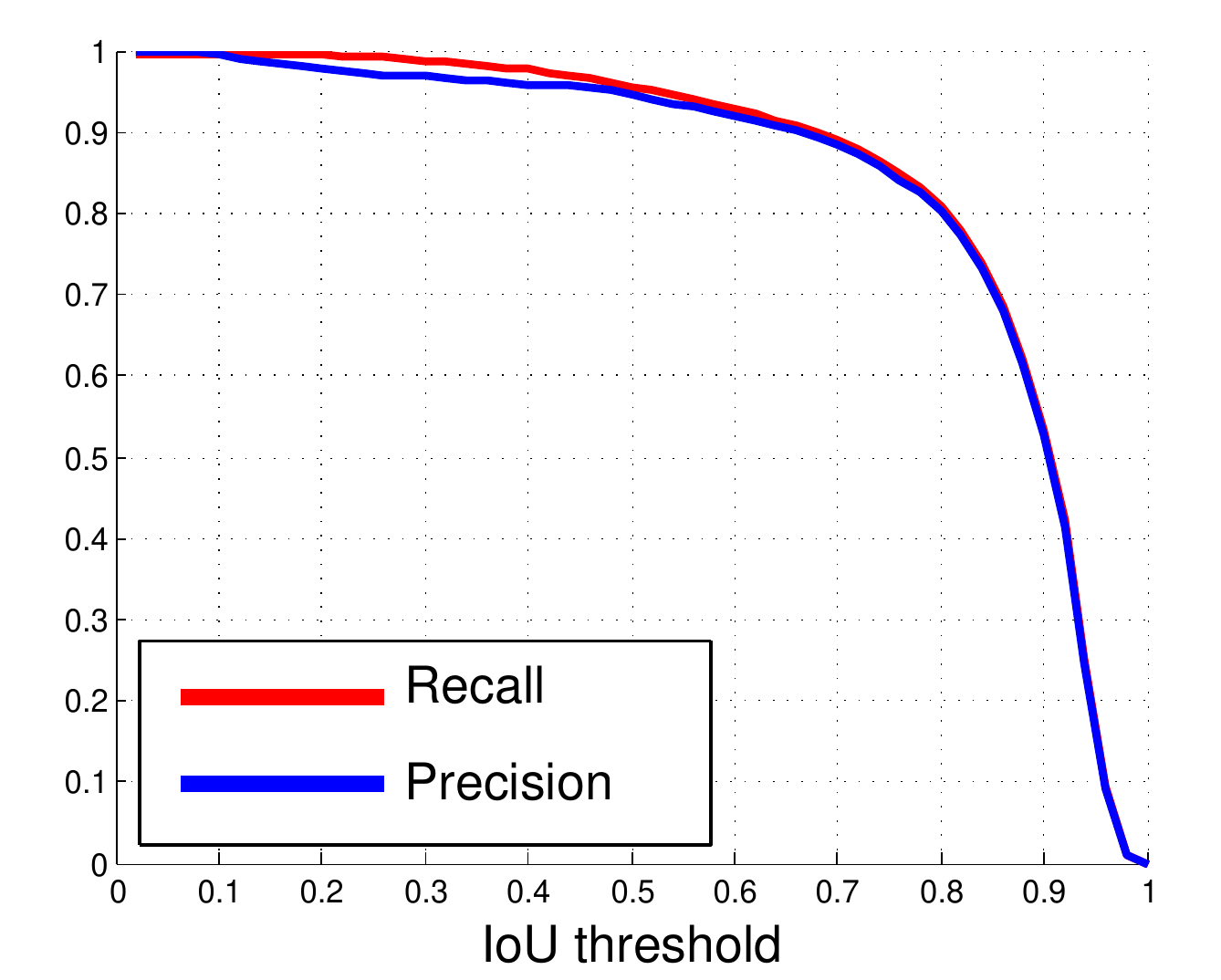}
	\caption{Recall and precision plots of the quantitative result.}
	\vspace{-15pt}
	\label{fig6}
\end{figure}

Figure~\ref{fig6} shows the recall and precision of detecting foreground instances with a changing IoU metric. Given a specific threshold value $IoU=0.5$, the corresponding recall and precision are listed in Table~\ref{tab1}. 

\begin{table*}[htbp]
	\caption{Quantitative result.}
	\label{tab1}
	\begin{center}
	\begin{tabular}{p{8mm}p{6mm}<{\centering}p{6mm}<{\centering}p{6mm}<{\centering}p{6mm}<{\centering}p{6mm}<{\centering}p{6mm}<{\centering}p{6mm}<{\centering}p{6mm}<{\centering}p{6mm}<{\centering}p{6mm}<{\centering}p{6mm}<{\centering}}
	\hline
	\specialrule{0em}{1pt}{2pt}
	\textbf{Name} & \textbf{1}&\textbf{2}&\textbf{3}&\textbf{4}&\textbf{5}&\textbf{6}&\textbf{7}&\textbf{8}&\textbf{9}&\textbf{10}&\textbf{Avg}\\
	\hline
	\specialrule{0em}{1pt}{2pt}
	\textbf{Re} & 0.973 & 0.851 & 0.989 & 0.949 & 0.997 & 0.977 & 0.980 & 0.938 & 0.925 & 0.971 & 0.955\\
	\hline
	\specialrule{0em}{1pt}{2pt}
	\textbf{Pr} & 0.954 & 0.950 & 0.896 & 0.992 & 0.988 & 0.925 & 0.953 & 0.944 & 0.933 & 0.926 & 0.946\\
	\hline
	\multicolumn{12}{p{120mm}}{Note: The ten sequences from 1 to 10 are: Backdoor, BusStation, CopyMachine, Cubicle, Office, Pedestrians, PeopleInShade, PETS2006, Skating, Sofa \cite{CDNet2014}.}
	\end{tabular}
	\end{center}
\end{table*}

\begin{figure*}[htbp]
\centering
\subfigure[Time consumption] { \label{fig:11a} 
\begin{minipage}[htb]{0.48\textwidth}
\includegraphics[scale=0.55]{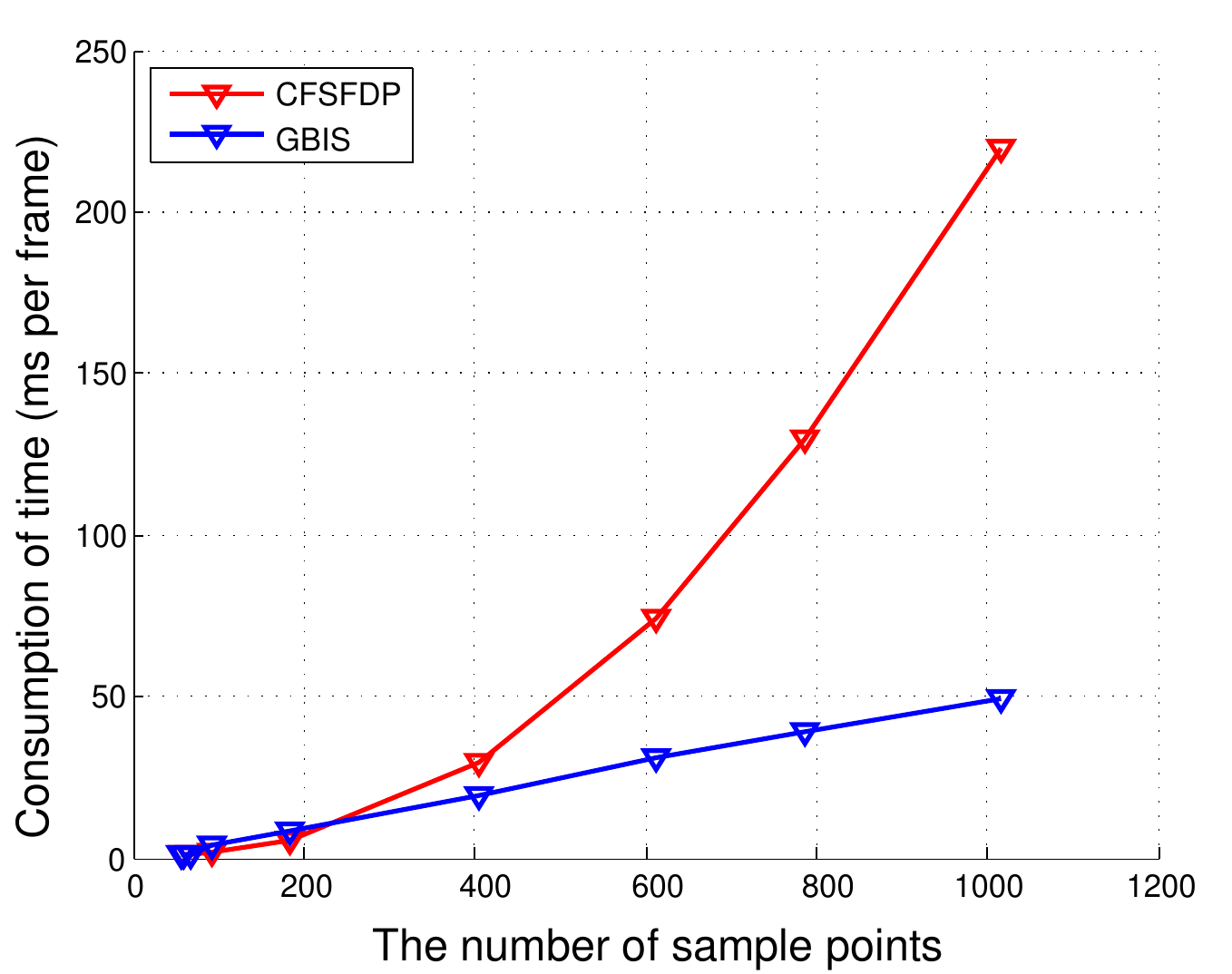}
\end{minipage}
}
\subfigure[Effect] { \label{fig:11b} 
\begin{minipage}[htb]{0.48\textwidth}
\includegraphics[scale=0.55]{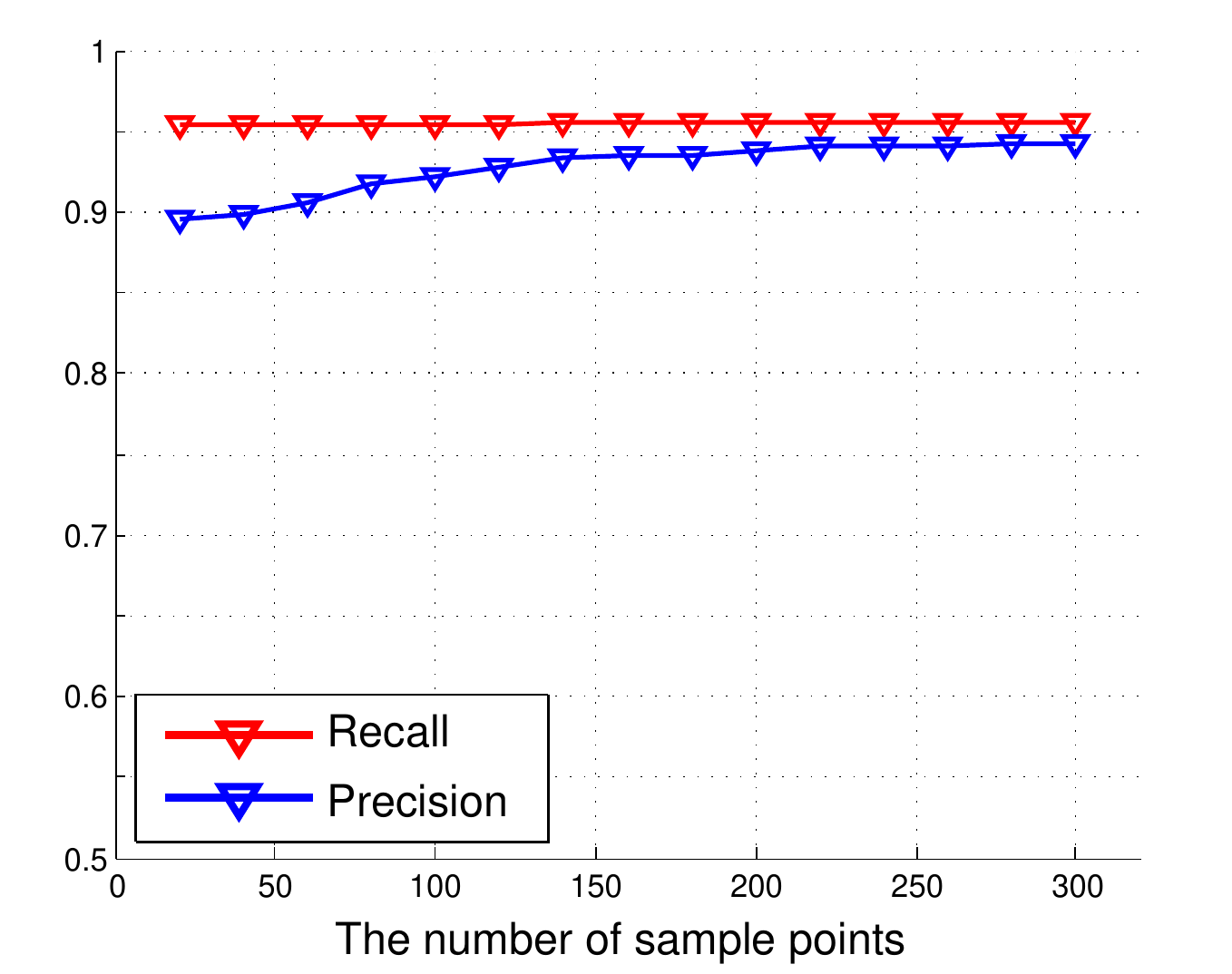}
\end{minipage}
}
\caption{Illustration of the effect of the CFSFDP method sample point number. From left to right: The time consumption curve of CFSFDP method and GBIS method related to the number of foreground sample points. The recall and precision result of ten video sequences related to the number of foreground sample points in CFSFDP method.}
\label{fig11}
\end{figure*}

\subsection{Qualitative result}\label{AA}
Some bounding box results obtained by our method are illustrated in Figure~\ref{fig4}. As the scenes contain different challenges, comparing the detecting result (the red bounding boxes) with the ground truth (the yellow bounding boxes), one can visually make out that the proposed method can output instance-level bounding boxes with a high success rate as well as high quality.

Obviously our method outputs result with high recall and high precision. An average recall exceeding ninety percent means that most of instances inside the scenes have been successfully found out, which is enough high for practical applications. On the other hand, an average precision over ninety percent means that we obtain a satisfying result at the cost of a very small amount of false positive results. This is of significance in practical applications as too much false positive results will cause huge extra computing consumption in the subsequent processes.


\subsection{Effect of composition analyzing}

Table~\ref{tab2} shows the contributions of CFSFDP method. Without using CFSFDP for composition analysis, the recall drops slightly. This is because the parameter $\tau$ is set as a constant, as the framework didn't have any message of how many objects inside the scenes. The foreground segmentation process lacks of a adaptive desired section size for guidance. As a result, some segments either only contain a section of an object or contain more than one object. Besides, the precision drops sharply. This is because without Formula~\eqref{eq14}, all segments produced by foreground segmentation process are considered as a protential objects. As shown in Figure~\ref{fig1}, $S$ contains many tiny false positive segments compared with $F^s$. Only through combining CFSFDP and GBIS, can the framework suppress most false positive results and greatly improve the precision performance.

\begin{table}[h]
  	\centering
  	\caption{The average recall and precision results for the contributions of CFSFDP method.}
  	\label{tab2}
  	\begin{tabular}{p{25mm}p{10mm}<{\centering}p{10mm}<{\centering}}
		\hline
		\specialrule{0em}{1pt}{2pt}
		\textbf{Method} & \textbf{Re}&\textbf{Pr}\\
		\hline
		\specialrule{0em}{1pt}{2pt}
		\textbf{GBIS} & 0.904 & 0.202 \\
		\hline
		\specialrule{0em}{1pt}{2pt}
		\textbf{CFSFDP+GBIS} & 0.955 & 0.946\\
		\hline
		\multicolumn{3}{p{60mm}}{Note: $IoU=0.5$.}
	\end{tabular}
\end{table}

\subsection{Efficiency}
Table~\ref{tab3} shows the computation time measured by Matlab on an Intel Core i5-7400 3.0GHz PC. As shown in the result, the optical flow estimating process and the foreground segmenting process occupy most of the total time consumption. The composition analysis process and the postprocessing process spend relatively less time. 

\begin{table}[h]
  	\centering
  	\caption{Time consumption of the proposed method.}
  	\label{tab3}
  	\begin{tabular}{p{40mm}p{10mm}<{\centering}}
		\hline
		\specialrule{0em}{1pt}{2pt}
		\textbf{Process} &\textbf{Time(ms)}\\
		\hline
		\specialrule{0em}{1pt}{2pt}
		\textbf{Optical flow extracting}&${123}^\star$ \\
		\specialrule{0em}{1pt}{2pt}
		\textbf{Composition analyzing}&7 \\
		\specialrule{0em}{1pt}{2pt}
		\textbf{Foreground segmenting}&87\\
		\specialrule{0em}{1pt}{2pt}
		\textbf{Postprocessing}&11\\
		\specialrule{0em}{1pt}{2pt}
		\textbf{Total} & 228 \\
		\hline
		\multicolumn{2}{p{60mm}}{Note: The entries show the time consumption of each process in the form ms per frame. $\star$result is quoted from \cite{FlowNet2}.}
	\end{tabular}
\end{table}

Further experiment shows that the time consumption of CFSFDP method is proportional to the square of the sample point number. And the time consumption of GBIS method represents linearity correlation to the number of sample points. Computational consumption of CFSFDP method is a bottleneck when the number of sample points increases as shown in Figure~\ref{fig:11a}. However, as shown in Figure~\ref{fig:11b}, the increase of sample point number contributes little to the detection results in fact, when the number of sample points is above 200. Thus, we set the sample point number as $N_c=200$.


\section{Conclusion}
\label{sec:CC}
In this work, we focus on the problem that analyzes the foreground masks obtained by moving object detection and outputs the instance-level moving object information. This is of great significance to the application of moving object detection. To address this problem, we proposed an optical flow based framework mainly utilizing two complementary clustering algorithms to analyze and segment the foreground. Beside, our frame output several kinds of moving objects information, which can be directly used in the following procedures like track or instance analysis. In experiment part, we use quantitative and qualitative results to indicate that our framework is designed properly and is effective enough for most practical applications.

\end{document}